\pdfoutput=1

\documentclass[11pt]{article}

\usepackage[preprint]{acl}

\usepackage{times}
\usepackage{latexsym}

\usepackage[T1]{fontenc}

\usepackage[utf8]{inputenc}

\usepackage{microtype}

\usepackage{inconsolata}

\usepackage{graphicx}
\usepackage{booktabs}
\usepackage{multirow}
\usepackage{amssymb}
\usepackage{amsmath}
\usepackage{amsthm}
\newtheorem{theorem}{Theorem}[section]
\newtheorem{definition}[theorem]{Definition}

\newtheorem{prof}{Proof}[section]
\title{Revisiting Self-Consistency from Dynamic Distributional Alignment Perspective on Answer Aggregation}

\author {
    \textbf{Yiwei Li}\textsuperscript{\rm 1}\footnotemark[1], \hspace{0cm}
    \textbf{Ji Zhang}\textsuperscript{\rm 1}\footnotemark[1], \hspace{0cm}
    \textbf{Shaoxiong Feng}\textsuperscript{\rm 2}, \hspace{0cm} 
    \textbf{Peiwen Yuan}\textsuperscript{\rm 1}, \hspace{0cm} 
    \textbf{Xinglin Wang}\textsuperscript{\rm 1}, \hspace{0cm} \\
    \textbf{Jiayi Shi}\textsuperscript{\rm 1}, \hspace{0cm} 
    \textbf{Yueqi Zhang}\textsuperscript{\rm 1}, \hspace{0cm} 
    \textbf{Chuyi Tan}\textsuperscript{\rm 1}, \hspace{0cm} 
    \textbf{Boyuan Pan}\textsuperscript{\rm 2}, \hspace{0cm} 
    \textbf{Yao Hu}\textsuperscript{\rm 2}\textbf{,} \hspace{0cm} 
    \textbf{Kan Li}\textsuperscript{\rm 1}\footnotemark[2] \\
    \textsuperscript{\rm 1} School of Computer Science, Beijing Institute of Technology \\
    \textsuperscript{\rm 2} Xiaohongshu Inc \\
    \texttt{\{liyiwei,jizhang,peiwenyuan,wangxinglin\}@bit.edu.cn} \\
    \texttt{\{shaoxiongfeng2023\}@gmail.com} \  \texttt{\{panboyuan,xiahou\}@xiaohongshu.com} \\
     \texttt{\{shijiayi,zhangyq,tanchuyi,likan\}@bit.edu.cn} 
}

\begin{document}
\maketitle

\renewcommand{\thefootnote}{\fnsymbol{footnote}} 
\footnotetext[1]{Equal contribution.} 
\footnotetext[2]{Corresponding author.} 

\renewcommand{\thefootnote}{\arabic{footnote}}

\begin{abstract}
Self-consistency improves reasoning by aggregating diverse stochastic samples, yet the dynamics behind its efficacy remain underexplored. We reframe self-consistency as a dynamic distributional alignment problem, revealing that decoding temperature not only governs sampling randomness but also actively shapes the latent answer distribution.  Given that high temperatures require prohibitively large sample sizes to stabilize, while low temperatures risk amplifying biases, we propose a confidence-driven mechanism that dynamically calibrates temperature: sharpening the sampling distribution under uncertainty to align with high-probability modes, and promoting exploration when confidence is high. Experiments on mathematical reasoning tasks show this approach outperforms fixed-diversity baselines under limited samples, improving both average and best-case performance across varying initial temperatures without additional data or modules. This establishes self-consistency as a synchronization challenge between sampling dynamics and evolving answer distributions.
\end{abstract}
\section{Introduction}
Self-consistency \citep{COT} is a well-established decoding method that enhances model performance by aggregating multiple stochastic samples via majority voting. It has been demonstrated to be highly effective across a variety of tasks \citep{USC,wang-etal-2024-integrate}, particularly in improving reasoning abilities \citep{COT}. Despite its empirical success, the underlying mechanisms behind self-consistency remain underexplored.
In this work, we revisit self-consistency from a distributional perspective, reframing it as a dynamic alignment problem, to achieve more robust and effective performance in answer aggregation. 

\begin{figure}[t]
\centering
\includegraphics[width=0.8\linewidth]{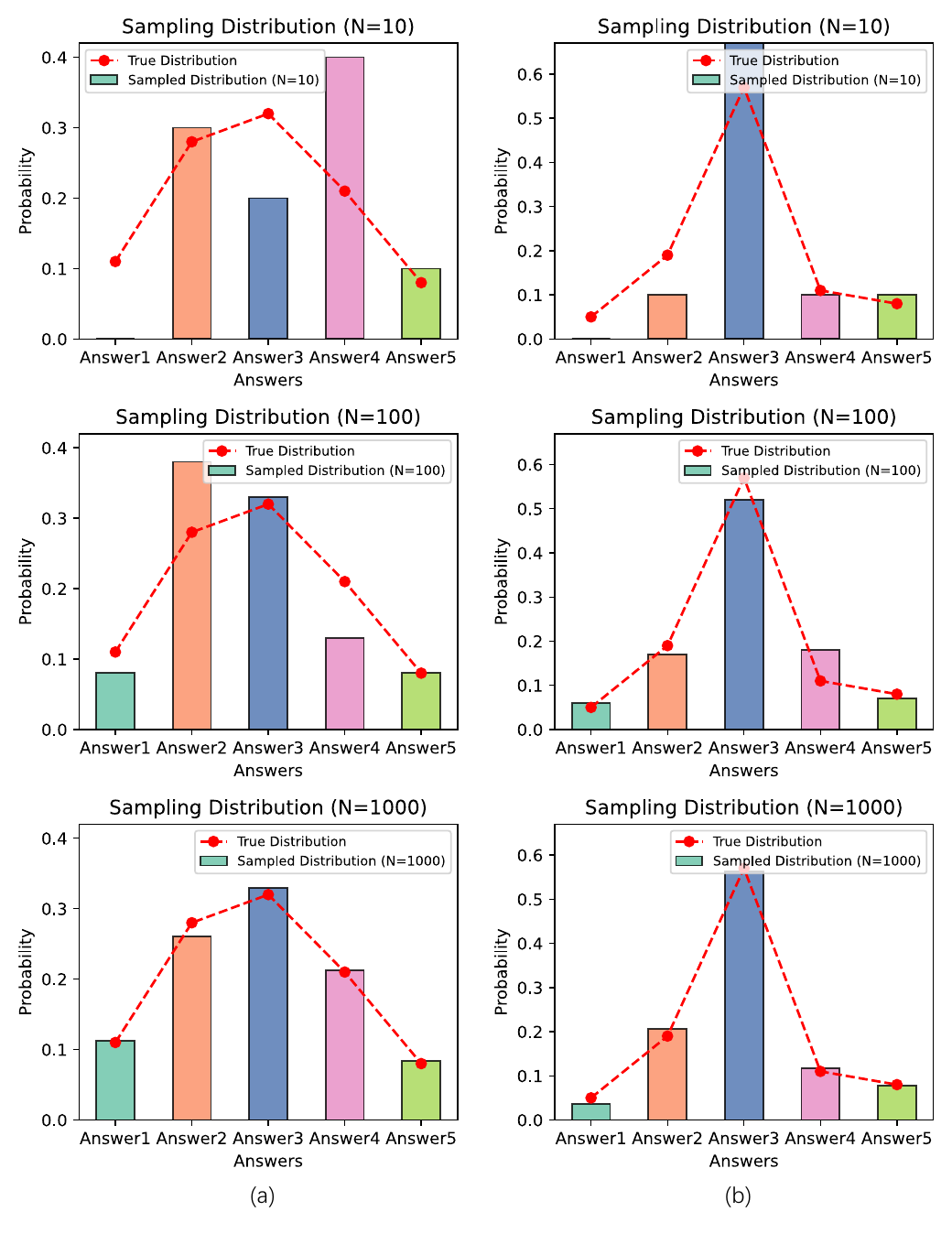}
\caption{(a) Multiple stochastic sampling for fitting the true distribution. As the sample size increases, the noise gradually diminishes, and ultimately, the top-1 sampled answer aligns with the true distribution. (b) As the temperature decreases, the confidence in the true distribution increases, allowing alignment with the true distribution to be achieved with fewer samples.}
\label{fig:intro}
\end{figure}

Recent work \citep{ISL,ESC} argue that by combining different reasoning traces via majority voting, self-consistency can avoid local optima and reduce the high variance associated with single-sample outputs, ultimately converging to the model’s true answer distribution (see Figure~\ref{fig:intro} (a)). Building on this insight, our work provides a formal definition of its convergence and derives practical criteria for the assessment. Through our convergence analysis, we reveal that this conventional view is limited to a fixed true distribution, overlooking the crucial impact that parameter-controlled decoding (typically the temperature) has on the true distribution (see Figure~\ref{fig:intro} (b)). Moreover, practical applications are often constrained by the sampling. Therefore, we raise two key questions:
(1) Alignment under Constraints: How does decoding diversity affect the alignment between the sampling distribution and the true answer distribution when only a limited number of samples is available? (2) Dynamic Alignment: Can we actively calibrate the diversity in practice to accelerate and stabilize convergence, rather than passively waiting for asymptotic convergence?

To explore these issues, we analyze the impact of diversity on self-consistency. The temperature parameter not only governs the randomness of sampling but also directly shapes the true answer distributions. Our findings reveal that as the number of samples approaches infinity, a higher temperature yields a more ideal true answer distribution. However, when the sample size is finite, the optimal sampling temperature decreases as the number of samples diminishes. This leads to a trade-off: low-diversity sampling quickly concentrates the answers and suppresses noise but risks amplifying model biases, whereas high-diversity sampling disperses the answers, requiring more samples to stabilize, yet it enables the exploration of a potentially superior true distribution.

In summary, our comprehensive analysis indicates that the effectiveness of self-consistency hinges on a dynamic alignment between the confidence of the sampling distribution and the intrinsic uncertainty of the true answer distribution—a relationship that is influenced by the number of samples. Ideally, the sampling distribution should be controlled such that the majority voting outcomes closely match the true distribution, and on this basis, explores toward an improved true distribution.

Based on this insight, we propose a confidence-driven diversity optimization mechanism that dynamically adjusts the temperature based on real-time confidence values derived from the answer distribution. When early samples show only a small probability gap between the top two most-voted answers, our mechanism sharps the sampling distribution to better align it with the true distribution. Conversely, when confidence is high, the temperature is increased to explore potentially superior distributions. We derive a confidence threshold to determine the direction of temperature adjustment, providing theoretical support for this process. This closed-loop control dynamically synchronizes the sampling distribution with the latent answer distribution, ensuring efficient convergence while actively pursuing a better distribution.
Experimental results across various model types and size indicate that it can achieve simultaneous improvements in both average and best performance across different initial temperatures, without requiring any additional training, valid data, reward models, or external modules.
\section{Fundamental Analysis of Self-Consistency}

In this section, we first present a distribution alignment perspective on how self-consistency works with specific true answer distributions, supported by experimental evidence to substantiate this viewpoint. Building upon this foundation, we proceed to provide both a formal definition of self-consistency convergence and practical criteria for assessment. 
\subsection{Why Self-Consistency Works: A Distributional Perspective}

Self-Consistency is a widely-used decoding method for improving reasoning performance by aggregating multiple stochastic samples. 
By applying a majority voting scheme, it mitigates issues such as local optima and high variance that arise from relying on a single sample. Formally, it can be expressed as:

\begin{equation}
    \hat{y}_{SC} = \arg\max_y \left( \frac{1}{n} \sum_{i=1}^{n} \mathbb{I}(y_i = y) \right)
\end{equation}
where $y_i$ is the $i$-th sampled answer, and $\mathbb{I}(y_i = y)$ is the indicator function that equals 1 if $y_i$ matches the candidate answer $y$, and 0 otherwise. The result, $\hat{y}_{SC}$, is the answer with the highest number of votes (the top-1 answer).

From a probabilistic perspective, self-consistency can be seen as a \textit{Monte Carlo estimator} of the true answer distribution $p(y \mid \mathbf{x})$. As the number of samples increases, the empirical distribution formed by the samples approximates the true distribution, and the most frequent answer aligns with the true distribution:
\begin{equation}
\begin{aligned}
\hat{p}_{SC}(y) &= \frac{1}{n} \sum_{i=1}^{n} \mathbb{I}(y_i = y) \\
&\to p(y \mid \mathbf{x}), \quad \text{as} \quad n \to \infty
\end{aligned}
\end{equation}
As the number of samples increases, the estimation becomes more reliable, and the voting mechanism converges towards the true answer.

\paragraph{Experimental Analysis}
\begin{figure}[ht]
\centering
\includegraphics[width=0.9\linewidth]{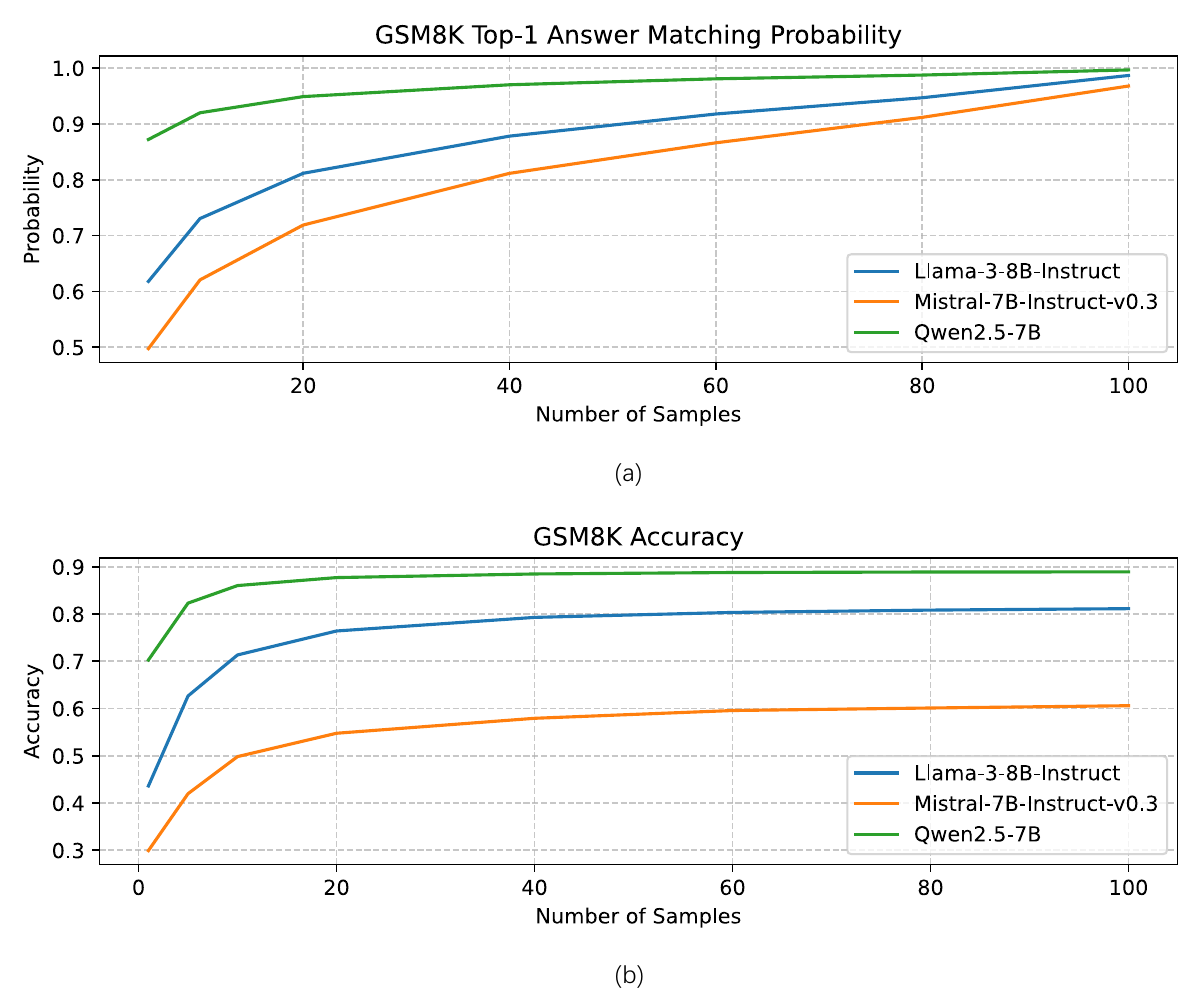}
\caption{Top-1 answer matching probability (a) and accuracy (b) both improve as the sampling number increases.}
\label{fig:top1}
\end{figure} 
To validate this viewpoint, we analyzed the top-1 answer match rate as a function of the sample size. The true top-1 answer is simulated by drawing from a large sample to approximate the true distribution. 
Results from Figure~\ref{fig:top1} reveals \textbf{Findings 1}: As the sample size increased, the top-1 answer match rate gradually approaches 100\% with the accuracy consistently improves.
Based the observation, we derive the following insight: 
\textit{\textbf{Insights 1}: The improvement in self-consistency performance stems from the fact that, 
the top-1 answer in the sampling distribution gradually aligns with the true distribution, ultimately enhancing accuracy to match the true distribution's level.}

\subsection{Convergence Analysis of Answer Aggregation}
According to \textit{\textbf{Insights 1}}, since the accuracy of the true distribution is fixed, the performance of self-consistency is guaranteed to converge.
To further investigate it, we provide the following definition according to the Cauchy convergence criterion:
\begin{definition}
Let $f^M(i) = \sum_{l=1}^M \mathbb{I}(\hat{y_l} = i)$, where $\hat{y_l}$ represents the set of answers generated by the model, and $ M $ is the number of samples. For any given $ \epsilon > 0 $, there exists a positive integer $ L $ such that for $ N, M > L $, if the following holds:
\begin{align}
\left|\; \underset{i}{argmax}\; f^M(i) - \underset{i}{argmax}\; f^N(i) \;\right| < \epsilon
\end{align}
we can conclude that self-consistency has converged.
\label{def:sc}
\end{definition}

Based on Definition \ref{def:sc}, we prove that self-consistency also converges in terms of the accuracy on the dataset:

\begin{theorem}
Let $ Acc_{D}^M = \frac{1}{|D|}\sum_{j\in D}\mathbb{I}[\underset{i}{argmax}\; f^M(i)=gt_j] $ denote the accuracy of self-consistency when a single question is sampled $ M $ times on dataset $ D $, where $ gt_j $ represents the correct answer to the $ j $-th question. If Definition 1 holds, then for any given $ \epsilon > 0 $, there exists a positive integer $ L $ such that when $ N, M > L $, the following holds:
\begin{align}
\left| \; Acc_{D}^M - Acc_{D}^N \;\right| < \epsilon
\end{align}
\label{the:sc}
\end{theorem}
The Proof of Theorem~\ref{the:sc} is in Appendix~\ref{app:proof}.
By setting $\epsilon$ to $\frac{1}{|D|}$, the following definition is established:
\begin{definition}
If the following holds on dataset $D$:
\begin{align}
\left| \; Acc_{D}^M - Acc_{D}^{M-5} \;\right| < \frac{1}{|D|}
\end{align}
we can consider self-consistency to have converged at a sample size of $M$.
 
\label{def:sc_D}
\end{definition}
\paragraph{Experimental Analysis}
\begin{figure}[t]
\centering
\includegraphics[width=1.0\linewidth]{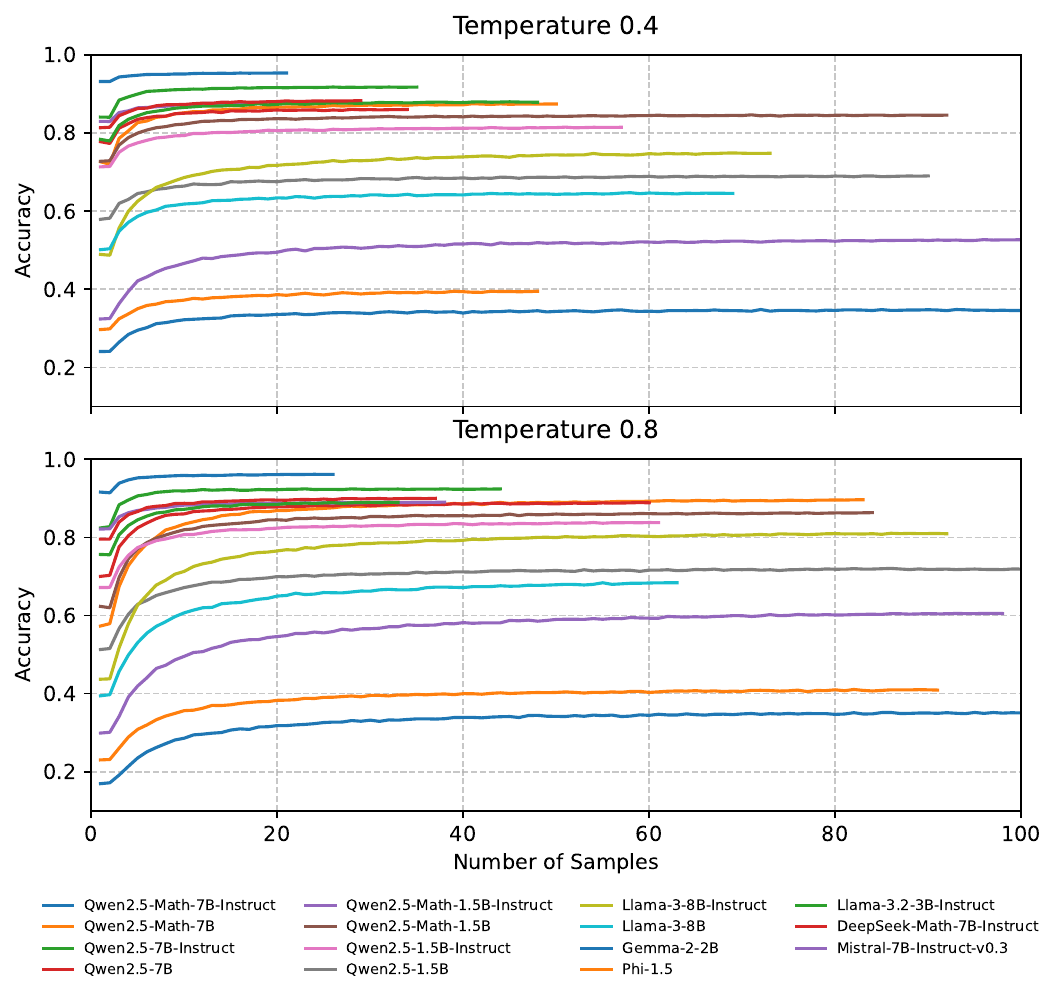}
\caption{Self-consistency convergence plots under different temperature (0.4 and 0.8) settings.}
\label{fig:convergence}
\end{figure} 
Figure~\ref{fig:convergence} depicts the convergence behavior of various models, with the accuracy curves plotted up to the convergence point according to Definition~\ref{def:sc_D}, from where we can get:
\textbf{Findings 2}: The convergence speed exhibits a positive correlation with accuracy.
\textbf{Findings 3}: The convergence speed is inversely correlated with temperature. 
\textbf{Findings 4}: The final converged accuracy varies across different temperature settings.
Based on them, we derive \textit{\textbf{Insights 2}: Sampling diversity will affect the true distribution, impacting both the convergence accuracy and the convergence speed of self-consistency.}

\section{Diversity Trade-offs for Self-Consistency}
\label{sec:diversity}
\subsection{Sampling Diversity Affection}
According to \textbf{\textit{Insight 2}}, to gain a deeper understanding of the impact of diversity on self-consistency, we investigate how accuracy varies with temperature changes in increments of 0.1. The study is divided into two parts: convergence analysis and finite-sample analysis.
\paragraph{Converge Condition}

\begin{figure}[t]
\centering
\includegraphics[width=0.75\linewidth]{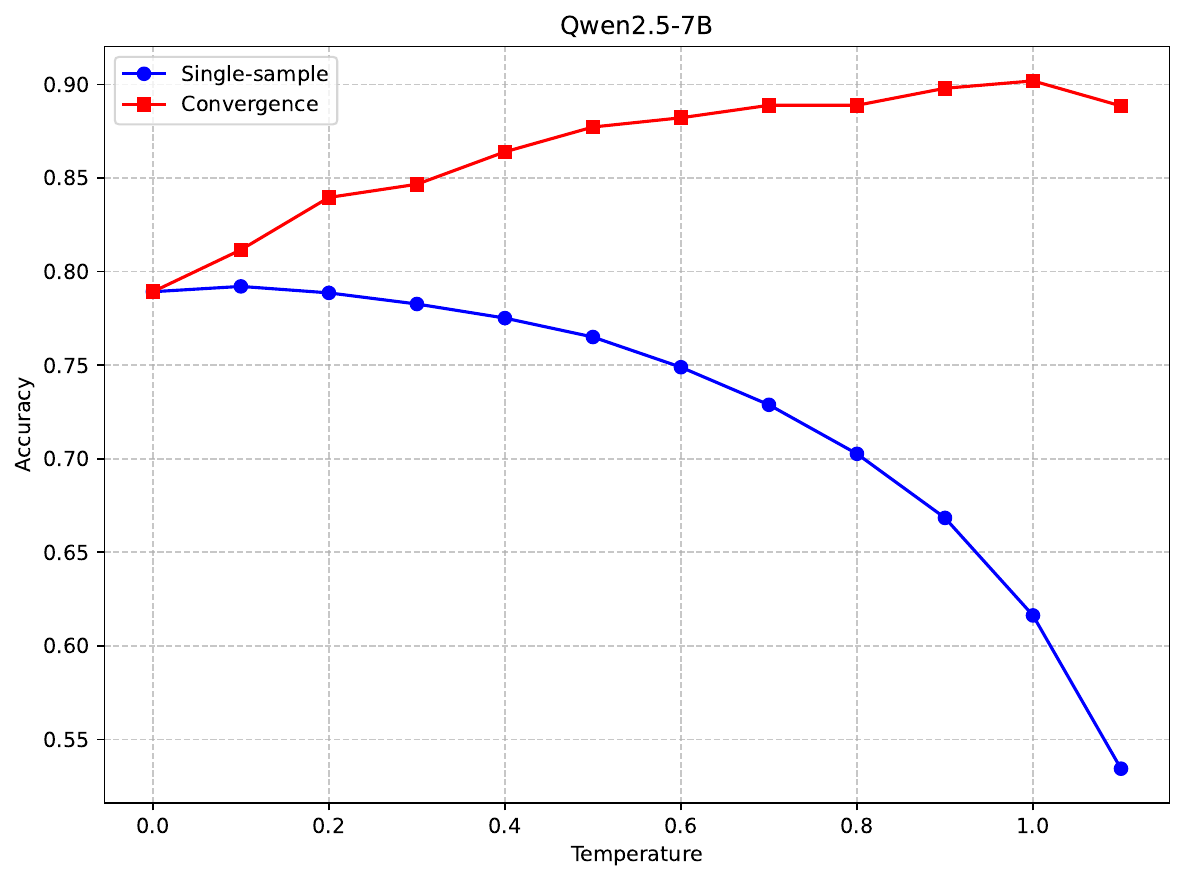}
\caption{The accuracy curve with varying temperature under convergence.}
\label{fig:inf_tem}
\end{figure} 
Figure~\ref{fig:inf_tem} indicates \textbf{Findings 5}: As the temperature increases, the accuracy of single samples exhibits a declining trend, while the accuracy of self-consistency after convergence shows an increasing trend (the optimal point is often near 1.0\footnote{We speculate that this may be related to the training temperature being typically set to 1.0. We leave the study of the optimal temperature as future work.}). Please refer to Appendix~\ref{app:diversity} for more results. The disagreement resolution theorem in ensemble learning provides a potential explanation, suggesting that the overall performance of an ensemble is determined by the trade-off between the accuracy of individual models and the diversity among them. From this trend and \textit{\textbf{Insights 1}}, we gain \textit{\textbf{Insights 3}: When the sample size is sufficient, the temperature should be increased to better explore the true distribution with higher accuracy.}

\paragraph{Finite-Sample Condition}

\begin{figure}[t]
\centering
\includegraphics[width=0.8\linewidth]{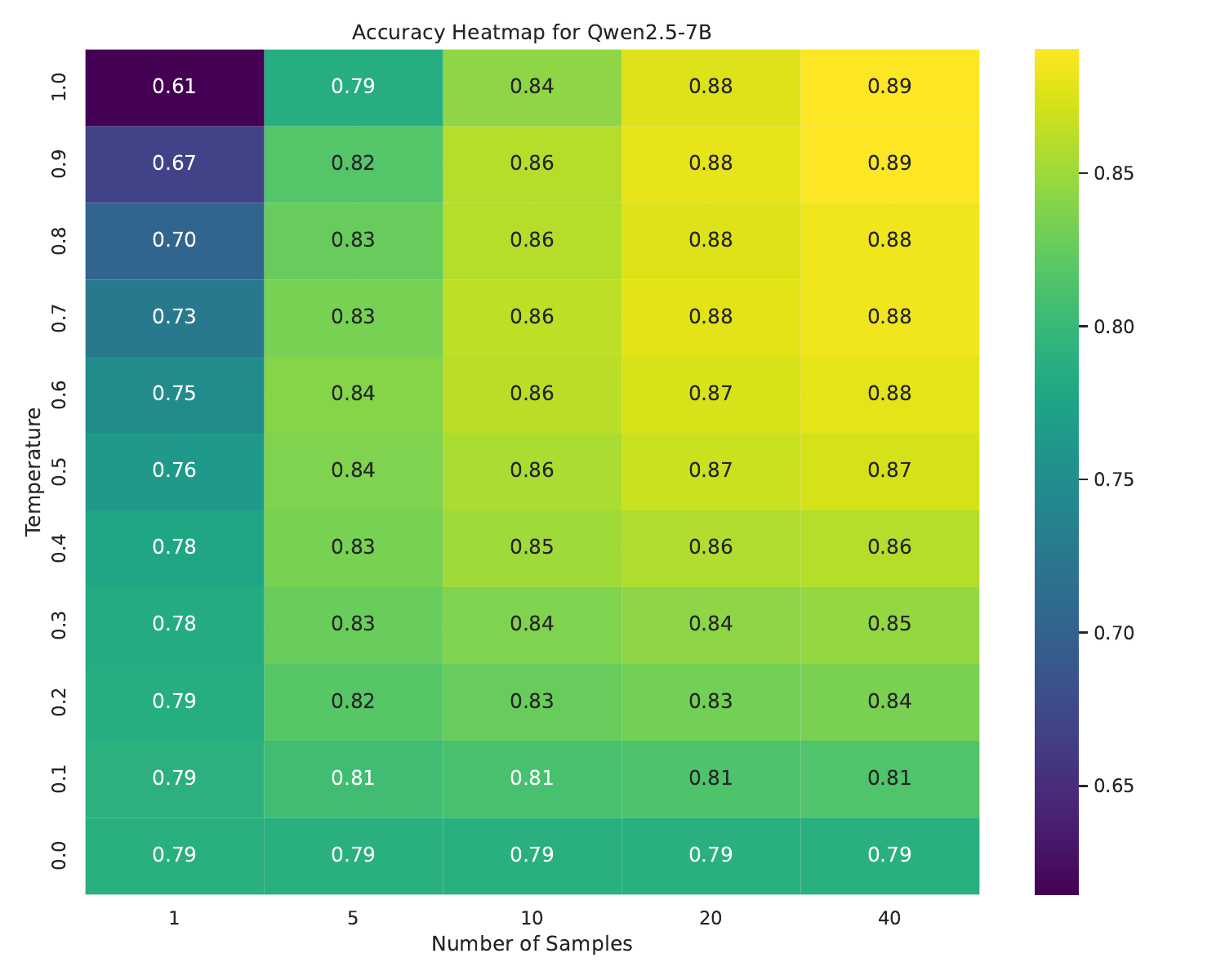}
\caption{The accuracy heatmap with varying temperature with finite sample size.}
\label{fig:limit_tem}
\end{figure} 
Figure~\ref{fig:limit_tem} indicates \textbf{Findings 6}: When the sample size is limited, the optimal temperature gradually shifts toward lower values as the sample size decreases.
Please refer to Appendix~\ref{app:diversity} for more results.
This findings and \textit{\textbf{Insights 1}} leads us to \textit{\textbf{Insights 4}: Sample size determines the maximum top-1 confidence level that can be reliably modeled. True distributions with lower confidence require larger data volumes to ensure that the sampled top-1 answer aligns with the converged result.}

By combining \textit{\textbf{Insights 3}} and \textit{\textbf{4}}, we can derive \textit{\textbf{Insights 5}: The effectiveness of self-consistency depends on dynamically aligning the confidence of the sampling distribution with the inherent uncertainty of the true answer distribution.}

\subsection{Chain-of-thought Affection}
\begin{figure}[t]
\centering
\includegraphics[width=0.8\linewidth]{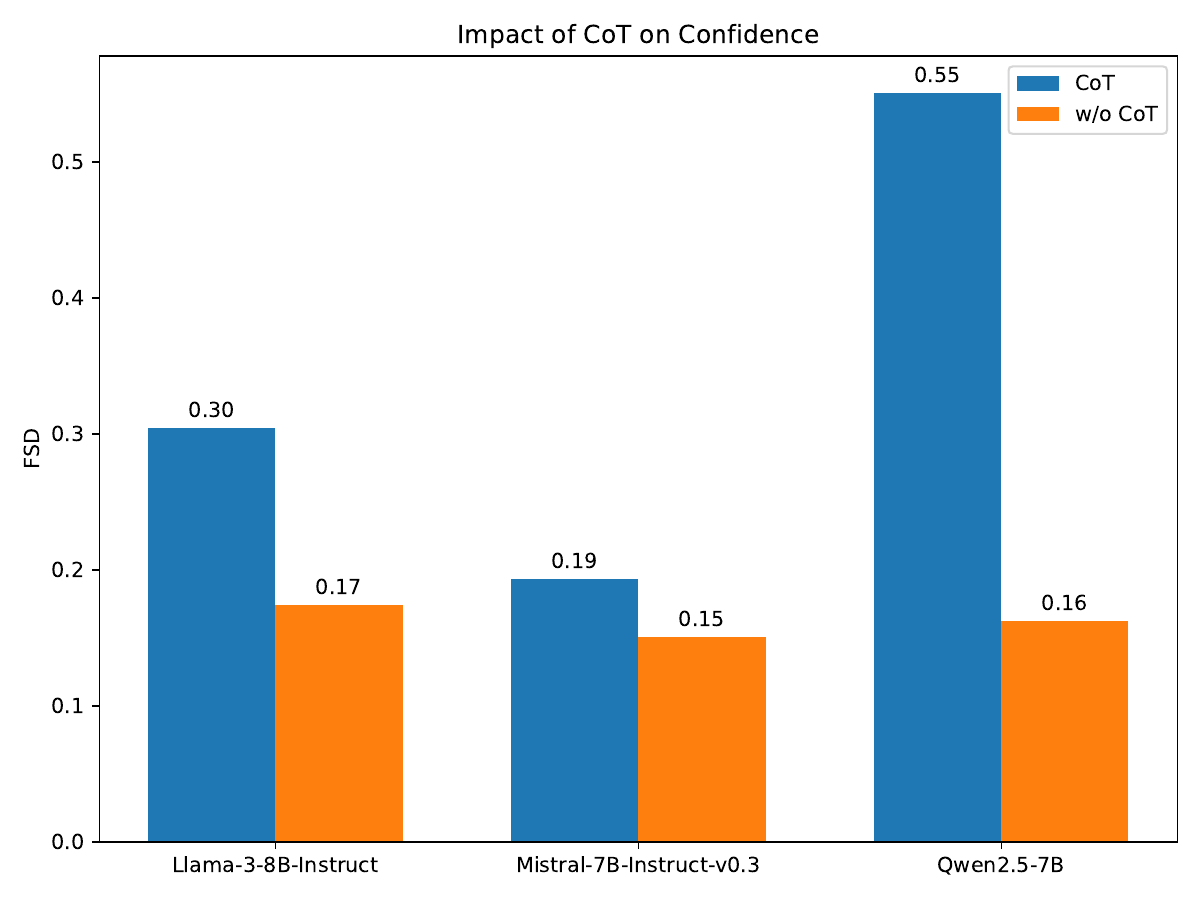}
\caption{FSD (Equation~\ref{eq:fsd}) \citep{FSD} is employed as the confidence metric to quantify the gap between top two candidates.}
\label{fig:cot_fsd}
\end{figure} 

Besides the sampling diversity decided by temperature, Chain-of-Thought \citep{COT} is also a key factor. From Figure~\ref{fig:cot_fsd} we can get \textbf{Findings 7}: Using CoT prompt leads to higher confidence compared to not using it.
A deeper \textbf{\textit{Insight 6}} emerges: \textit{Chain-of-thought (CoT) reasoning narrows the output space and reduces diversity, thereby increasing answer confidence.} However, investigating this phenomenon is not the focus of this paper, and we leave it for future work.

\section{From Static to Adaptive: Confidence-Driven Optimization of Self-Consistency Distributions}
\subsection{Motivation}
Traditional self-consistency methods rely on static sampling strategies with fixed sample sizes and temperatures, which face limitations:
When the confidence of the sampling distribution is too low, a limited sample size struggles to accurately capture the true top-1 answer, restricting the performance gains of self-consistency (\textbf{\textit{Insight 4}}). Conversely, when the confidence is too high, the model fails to explore potentially better distributions at higher temperatures (\textbf{\textit{Insight 3}}).
Our method addresses these issues through adaptive confidence-distribution alignment (\textbf{\textit{Insight 5}}). By dynamically adjusting the sampling distribution's diversity based on real-time confidence levels, we optimize alignment by proactively adapting to the evolving gap between model confidence and true distribution uncertainty. This dynamic mechanism enables efficient convergence to the correct answer even under limited sample sizes while facilitating exploration of better true distributions when needed. Through this approach, we enhance both the accuracy and robustness of self-consistency across diverse conditions.
\subsection{Diversity Control Strategy}

\paragraph{Dynamic Temperature Adjustment}
We introduce a confidence-driven diversity optimization mechanism to dynamically align the sampling distribution with the latent answer distribution. First-Second Distance (FSD) \citep{FSD} is employed as the confidence metric to quantify the gap between top candidates. Formally, at decoding step $t$:

\begin{equation}
     \text{FSD}^{(t)} = {p_1}^{(t)} - {p_2}^{(t)}
\end{equation}
where ${p_1}^{(t)}$ and ${p_2}^{(t)}$ are the probabilities of the top two answers from the first $t$ samples. This metric directly reflects the model’s uncertainty in distinguishing between the dominant candidates.

To ensure stable optimization, we design a conservative adjustment rule with a dead zone around confidence threshold $\tau$.
The temperature $T$ is updated based on the FSD metric:
\begin{align}
    T^{(t+1)} =
\begin{cases}
T^{(t)} - 0.1 & \text{if } \text{FSD}^{(t)} < \tau - \epsilon, \\
T^{(t)} + 0.1 & \text{if } \text{FSD}^{(t)} > \tau + \epsilon, \\
T^{(t)} &  \text{otherwise,}
\end{cases}
\label{eq:fsd}
\end{align}
where $\epsilon$ is a stability margin, which we set to 0.05 for simplicity. Temperature $T$ is clamped to [0.1, 1.0] to avoid extreme values.

\paragraph{Phased Sampling Strategy}
To balance exploration and efficiency, we employ a three-phase sampling protocol:
\begin{itemize}
    \item Exploration Phase: Collect small number of samples ($n_1 = 5$) with preset $T^{(1)}$ as a window to estimate initial $\text{FSD}^{(1)}$.
    \item Adaptive Phase: Adjust $T^{(2)}$ through Equation~\ref{eq:fsd}, then generate $n_2 = 0.5N - n_1$ ($N$: total budget) additional samples.
    \item Exploitation Phase: Finalize $T^{(3)}$ and generate the remaining $n_3 = 0.5 N$ samples.
\end{itemize}
The phased approach progressively shifts from broad exploration to focused exploitation. Finally, the accuracy is calculated by majority voting from the total of $N$ samples. 

In summary, our method dynamically adjusts the sampling diversity by monitoring the confidence levels, allowing for more efficient exploration and convergence. This adaptive mechanism ensures better alignment with the true answer distribution under sampling constraints to improve accuracy.

\subsection{Theoretical Analysis}
\label{sec:threshold}

To ensure a rational and effective selection of the FSD threshold $\tau$, we construct a one-sided z-test for analysis. The test employs the null hypothesis as follows:

$H_0$: The current sampled top-1 answer is not the true answer for the given question under infinite sampling.

To simplify this problem, we assume that only the current top-2 answer could potentially become the true answer under infinite sampling. Consequently, it is natural to focus on the relationship between FSD and confidence. Therefore, this hypothesis can be described as:
\begin{equation}
    z = \frac{ \hat{d} - d_{\mu} }{ SE }
\end{equation}
 where $\hat{d}$ represents the observed FSD from actual sampling, $d_{\mu}$ denotes the FSD under the true distribution of the model, and $SE$ stands for the standard error. Based on the null hypothesis $H_0$, it is clear that $d_{\mu}<0$. We adopt the critical condition $d_{\mu}=0$:

\begin{equation}
    z \geq \frac{ \hat{d} - 0}{ SE } = \frac{ \hat{p}_1 - \hat{p}_2 }{ SE }
\label{eq:z}
\end{equation}

Assuming that the current sample size approaches infinity and that the sampling between the two categories can be considered independent, according to the multinomial distribution and Jensen's inequality (in the case of a concave function), we have:
\begin{align}
    SE &=\sqrt{\frac{\hat{p}_1(1-\hat{p}_1)}{N}+\frac{\hat{p}_2(1-\hat{p}_2)}{N}} \notag\\
    &\leq \sqrt{\frac{2p(1-p)}{N}} 
\label{eq:SE}
\end{align}
where $p = \frac{\hat{p}_1+\hat{p}_2}{2} \in (0,0.5]$, substituting Equation \ref{eq:SE} into Equation \ref{eq:z}, we can derive the theoretical lower bound of $z$:
\begin{equation}
z \geq \frac{\hat{p}_1-\hat{p}_2}{\sqrt{\frac{2p(1-p)}{N}}} \geq \hat{d}\sqrt{2N}
\end{equation}

Setting $z = 1.64$, the corresponding $p$-value is approximately 0.05, indicating strong statistical evidence that the current top-1 answer is indeed the top-1 answer under the true model distribution. Therefore, the FSD threshold can be set as:
\begin{align}
    \tau = \frac{1.16}{\sqrt{N}}
\end{align}
\section{Experiments}
\begin{table*}[h]
    \centering
    \small
    \renewcommand\tabcolsep{3.5pt}
    \begin{tabular}{l l c c c c c c c c c c c c }
    \toprule
    \multirow{3}{*}{Models} & \multirow{3}{*}{Strategy} & \multicolumn{6}{c}{GSM8K} & \multicolumn{6}{c}{MATH} \\
    & & \multicolumn{2}{c}{N=10} & \multicolumn{2}{c}{N=20} & \multicolumn{2}{c}{N=40} & \multicolumn{2}{c}{N=10} & \multicolumn{2}{c}{N=20} & \multicolumn{2}{c}{N=40}\\ 
    \cmidrule(lr){3-4} \cmidrule(lr){5-6} \cmidrule(lr){7-8} \cmidrule(lr){9-10} \cmidrule(lr){11-12} \cmidrule(lr){13-14}
     & & Mean & Max & Mean & Max & Mean & Max & Mean & Max & Mean & Max & Mean & Max \\
     \toprule

\multirow{2}{*}{Qwen2.5-1.5B} & Fix & 65.4 & 67.6& 67.2 & 69.6& 68.2 & \textbf{70.9}& 31.4 & \textbf{36.1}& 34.5 & 38.6& 36.5 & 40.8 \\
& Dynamic & \textbf{65.7} & \textbf{67.7}& \textbf{67.8} & \textbf{69.8}& \textbf{68.9} & \textbf{70.9}& \textbf{32.4} & 36.0 & \textbf{36.5} & \textbf{38.9}& \textbf{38.7} & \textbf{41.0} \\
\midrule
\multirow{2}{*}{Qwen2.5-1.5B-Instruct} & Fix & 79.0 & \textbf{80.7}& 80.3 & 82.2& 81.1 & 83.2& 51.9 & \textbf{53.8}& 53.3 & 55.0& 54.1 & 55.8 \\
& Dynamic & \textbf{79.2} & \textbf{80.7}& \textbf{80.8} & \textbf{82.4}& \textbf{81.6} & \textbf{83.5}& \textbf{52.3} & 53.6 & \textbf{53.9} & \textbf{55.2}& \textbf{54.6} & \textbf{55.9} \\
\midrule
\multirow{2}{*}{Qwen2.5-7B} & Fix & 84.6 & 86.1& 85.7 & 87.7& 86.3 & 88.9& 48.7 & 52.0& 50.7 & 53.8& 51.8 & 54.9 \\
& Dynamic & \textbf{84.7} & \textbf{86.3}& \textbf{86.1} & \textbf{88.1}& \textbf{86.8} & \textbf{89.0}& \textbf{49.6} & \textbf{52.3}& \textbf{51.9} & \textbf{53.9}& \textbf{53.2} & \textbf{55.1} \\
\midrule
\multirow{2}{*}{Qwen2.5-7B-Instruct} & Fix & \textbf{90.8} & 91.9& 91.2 & 92.2& 91.4 & \textbf{92.4}& 65.9 & 66.6& 66.6 & \textbf{67.3}& 66.9 & \textbf{67.7} \\
& Dynamic & \textbf{90.8} & \textbf{92.0}& \textbf{91.4} & \textbf{92.3}& \textbf{91.7} & \textbf{92.4}& \textbf{66.1} & \textbf{66.7}& \textbf{66.8} & \textbf{67.3}& \textbf{67.2} & 67.6 \\
\midrule
\multirow{2}{*}{Qwen2.5-Math-1.5B} & Fix & 80.1 & \textbf{83.3}& 82.0 & 84.6& 82.9 & \textbf{85.6}& 41.5 & 44.1& 43.1 & \textbf{46.0}& 44.2 & 47.1 \\
& Dynamic & \textbf{80.7} & 83.2 & \textbf{82.9} & \textbf{84.7}& \textbf{83.9} & \textbf{85.6}& \textbf{41.9} & \textbf{44.2}& \textbf{44.1} & \textbf{46.0}& \textbf{45.2} & \textbf{47.2} \\
\midrule
\multirow{2}{*}{Qwen2.5-Math-1.5B-Instruct} & Fix & 87.1 & 88.2& 87.7 & \textbf{88.8}& 87.9 & 89.0& 64.2 & 65.1& 64.7 & \textbf{65.8}& 64.9 & \textbf{66.1} \\
& Dynamic & \textbf{87.2} & \textbf{88.4}& \textbf{87.9} & \textbf{88.8}& \textbf{88.2} & \textbf{89.2}& \textbf{64.3} & \textbf{65.2}& \textbf{64.8} & 65.6 & \textbf{65.1} & 65.9 \\
\midrule
\multirow{2}{*}{Qwen2.5-Math-7B} & Fix & 82.2 & 85.0& 84.6 & 87.1& 85.8 & 88.2& 52.7 & 56.1& 54.9 & 57.9& 56.3 & 59.4 \\
& Dynamic & \textbf{83.0} & \textbf{85.4}& \textbf{85.5} & \textbf{87.4}& \textbf{86.8} & \textbf{88.5}& \textbf{53.4} & \textbf{56.2}& \textbf{56.2} & \textbf{58.4}& \textbf{57.7} & \textbf{59.7} \\
\midrule
\multirow{2}{*}{Qwen2.5-Math-7B-Instruct} & Fix & 94.9 & 95.8& 95.2 & \textbf{96.0}& 95.4 & \textbf{96.2}& 68.8 & 70.1& 69.5 & \textbf{70.9}& 69.7 & \textbf{70.9} \\
& Dynamic & \textbf{95.1} & \textbf{95.9}& \textbf{95.4} & \textbf{96.0}& \textbf{95.6} & \textbf{96.2}& \textbf{69.3} & \textbf{70.4}& \textbf{70.0} & 70.7 & \textbf{70.2} & \textbf{70.9} \\
\midrule
\multirow{2}{*}{Llama-3-8B} & Fix & 58.2 & 63.0& 60.9 & 65.8& 62.5 & 67.4& 18.6 & 21.5& 20.3 & 23.5& 21.7 & 25.1 \\
& Dynamic & \textbf{59.3} & \textbf{63.4}& \textbf{62.6} & \textbf{66.1}& \textbf{64.3} & \textbf{67.6}& \textbf{19.3} & \textbf{21.9}& \textbf{22.1} & \textbf{24.2}& \textbf{23.6} & \textbf{25.5} \\
\midrule
\multirow{2}{*}{Llama-3-8B-Instruct} & Fix & 66.6 & 72.0& 70.2 & 76.1& 72.2 & 78.6& \textbf{20.1} & 24.4& 21.3 & 26.8& 22.1 & 28.7 \\
& Dynamic & \textbf{67.1} & \textbf{72.7}& \textbf{71.6} & \textbf{76.9}& \textbf{74.1} & \textbf{79.5}& \textbf{20.1} & \textbf{25.0}& \textbf{21.4} & \textbf{26.9}& \textbf{22.3} & \textbf{28.8} \\
\midrule
\multirow{2}{*}{Gemma-2-2B} & Fix & 29.1 & 32.2& 31.0 & 33.9& 32.3 & \textbf{34.9}& 14.6 & 16.5& 16.1 & \textbf{18.2}& 16.8 & 18.2 \\
& Dynamic & \textbf{29.7} & \textbf{32.3}& \textbf{32.3} & \textbf{34.2}& \textbf{33.5} & 34.7 & \textbf{15.1} & \textbf{16.7}& \textbf{16.8} & 17.9 & \textbf{17.6} & \textbf{18.4} \\
\midrule

\multirow{2}{*}{Phi-1.5} & Fix & 35.1 & 37.6& 37.0 & 39.5& 38.1 & 40.7& 4.0 & 4.7& 4.6 & 5.0& 5.0 & 5.5 \\
& Dynamic & \textbf{35.6} & \textbf{37.7}& \textbf{37.8} & \textbf{39.6}& \textbf{39.0} & \textbf{40.8}& \textbf{4.2} & \textbf{5.0}& \textbf{4.8} & \textbf{5.2}& \textbf{5.3} & \textbf{5.7} \\
\midrule
\multirow{2}{*}{DeepSeek-Math-7B-Instruct} & Fix & \textbf{87.4} & \textbf{88.6}& 88.1 & 89.5& 88.5 & \textbf{90.1}& 44.4 & 45.8& 46.2 & \textbf{48.2}& 47.1 & \textbf{49.5} \\
& Dynamic & \textbf{87.4} & \textbf{88.6}& \textbf{88.3} & \textbf{89.8}& \textbf{88.7} & \textbf{90.1}& \textbf{44.8} & \textbf{46.1}& \textbf{46.6} & \textbf{48.2}& \textbf{47.8} & \textbf{49.5} \\
\midrule
\multirow{2}{*}{Llama-3.2-3B-Instruct} & Fix & \textbf{86.2} & \textbf{87.4}& 87.2 & 88.4& 87.7 & 88.9& 48.7 & 49.8& 50.2 & 51.4& 51.2 & 52.4 \\
& Dynamic & \textbf{86.2} & 87.3& \textbf{87.5} & \textbf{88.6}& \textbf{88.1} & \textbf{89.2}& \textbf{49.0} & \textbf{50.1}& \textbf{50.6} & \textbf{51.6}& \textbf{51.7} & \textbf{52.7} \\
\midrule
\multirow{2}{*}{Mistral-7B-Instruct-v0.3} & Fix & 46.1 & 48.9& 49.3 & 53.3& 51.5 & 57.2& 17.1 & 18.3& 19.2 & 20.8& 20.8 & 22.4 \\
& Dynamic & \textbf{46.6} & \textbf{49.7}& \textbf{50.4} & \textbf{55.0}& \textbf{52.8} & \textbf{58.9}& \textbf{17.6} & \textbf{19.0}& \textbf{20.2} & \textbf{21.0}& \textbf{22.2} & \textbf{23.5} \\

    \bottomrule
    \end{tabular}
        
    \caption{Evaluation results by using 15 models from different base architectures on GSM8K\citep{GSM8K} and MATH\citep{MATH}. Dynamic temperature sampling achieves superior average and maximum performance across a wide range of settings.}
    \label{tb:method_results}
\end{table*}

\subsection{Experiment Setup}
\paragraph{Datasets and Models}
We evaluate our method on two widely-used mathematical reasoning benchmarks: GSM8K \citep{GSM8K} and MATH \citep{MATH}. 
Experiments span multiple model families to assess generalizability, including Qwen \citep{qwen}, Llama \citep{llama}, Mistral\citep{mistral}, DeepSeek\citep{deepseek}, Gemma\citep{gemma} and Phi\citep{phi}.
\paragraph{Implementation Details}
To systematically compare dynamic versus static temperature strategies, we test initial temperatures $T_{0}\in\{0.1,0.2,...,1.0\}$ with sampling budgets $N\in\{10, 20, 40\}$.
\paragraph{Metric}
To provide an intuitive and efficient evaluation of the differences between methods, we calculate both the average and maximum accuracy for fixed-temperature sampling and dynamic-temperature sampling across all temperatures. The evaluation is conducted from the perspectives of robustness and effectiveness. Formally:
\begin{align}
    Mean &= \frac{1}{N_T}\sum_{t \in T_0}Acc_t \\
    Max &= \underset{t \in T_0}{max} \;Acc_t 
\end{align}

\subsection{Results}
\label{sec:results}
From the results presented in Table \ref{tb:method_results} through 15 models, we can find:
\paragraph{Dynamic temperature sampling mitigates the performance degradation associated with fixed-temperature sampling.}
We find that the average accuracy across different temperatures for dynamic temperature sampling outperforms fixed-temperature sampling in the majority of models. This suggests that our method is not constrained by the temperature range and can identify samples that are more effective for self-consistency performance across various temperatures. To some extent, this approach mitigates the performance loss caused by ineffective sampling at a single fixed temperature.
\paragraph{For different samples, dynamic temperature sampling searches for a more suitable temperature for each sample.}
Similarly, we observe that dynamic temperature sampling also provides a certain improvement in terms of the maximum accuracy. This can be attributed to the fact that different samples require different temperature ranges. Fixed-temperature sampling can only achieve the desired accuracy for the dataset as a whole, whereas dynamic temperature sampling automatically searches for a more optimal temperature for each individual sample, maximizing the performance of self-consistency optimization across various temperatures.
\subsection{Analysis}
\label{sec:analysis}
\begin{figure*}[t]
\centering
\includegraphics[width=1.0\linewidth]{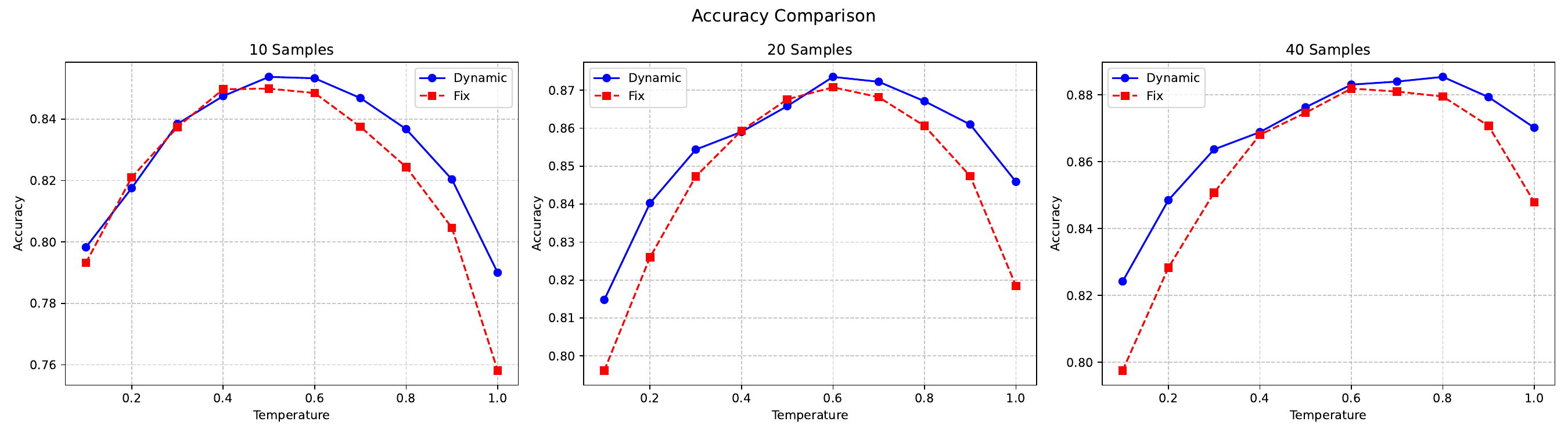}
\caption{A detailed results of the model's accuracy across different temperatures. Our method achieves better performance under both lower and higher initial temperatures.}
\label{fig:analysis_acc}
\end{figure*}

\paragraph{Visualization}
We provide a detailed analysis of the model's accuracy at different temperatures. Figure \ref{fig:analysis_acc} presents the accuracy and temperature curve for the Qwen2.5-Math-7B model. We observe that, with sampling sizes of 20 and 40, both low temperature ranges (0.1-0.4) and high temperature ranges (0.7-1.0) exhibit notable improvements. This suggests that dynamic temperature sampling yields more robust results. However, with a sampling size of 10, the performance in the low temperature range is almost identical to that of fixed-temperature sampling, primarily due to the constraints of the sample size. In the more optimal temperature range (0.4-0.7), the performance of dynamic and fixed-temperature sampling is similar, which aligns with our expectations and indicates that 
the intermediate temperature has already achieved a balanced trade-off.
\begin{figure*}[t]
\centering
\includegraphics[width=1.0\linewidth]{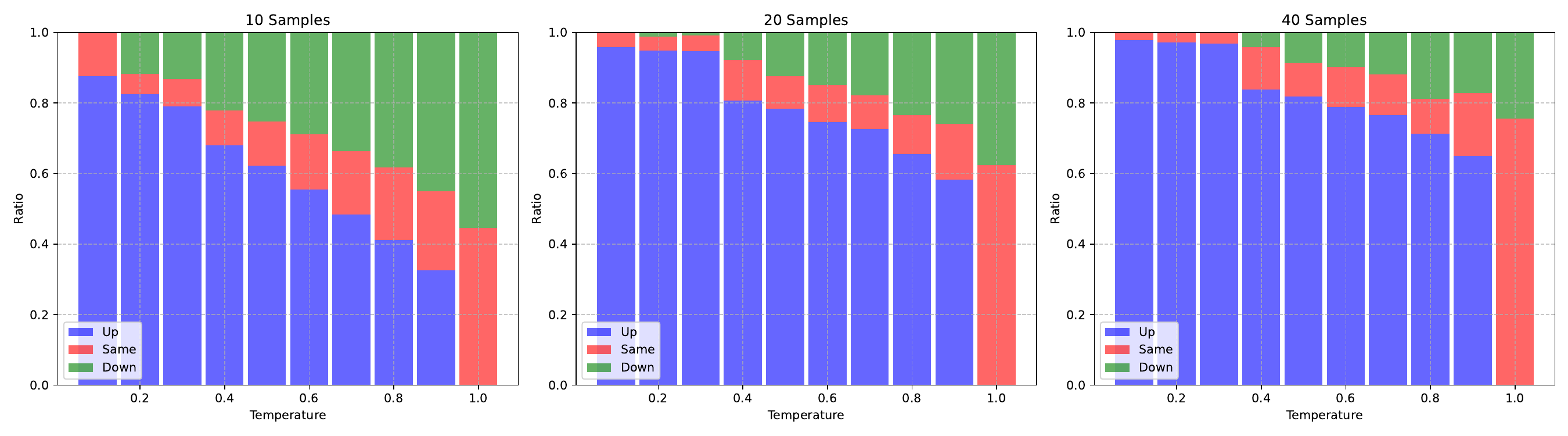}
\caption{Proportions of samples with temperature increases, decreases, or stability during dynamic temperature sampling.}
\label{fig:analysis_tem_ratio}
\end{figure*} 
\begin{figure*}[!ht]
\centering
\includegraphics[width=1.0\linewidth]{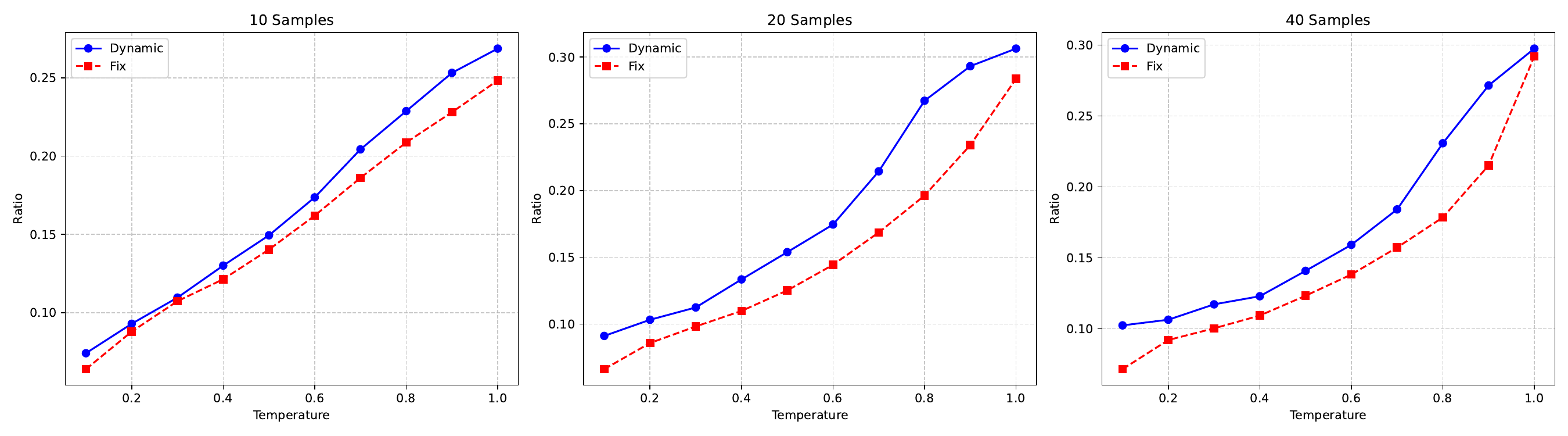}
\caption{Proportion of FSD instances reaching the dead zone, where dynamic temperature sampling results in a higher proportion.}
\label{fig:analysis_fsd_ratio}
\end{figure*} 
\paragraph{Direction Analysis of Temperature Variation}
Taking the sample level into account, we first analyzed the proportions of samples that experienced temperature increases, decreases, or remained constant throughout the dynamic temperature sampling process, as illustrated in Figure \ref{fig:analysis_tem_ratio}. We observed that in the low temperature range, at least 80\% of the samples experienced an increase in temperature. This observation is consistent with our hypothesis derived from dataset-level considerations, which suggests that increasing the temperature tends to result in higher expected accuracies. As the temperature rises, the proportion of samples experiencing temperature increases gradually declines, indicating that for some samples at the current sampling size, excessive temperatures are insufficient to confidently select the correct answer. Consequently, lowering the temperature becomes necessary to enhance FSD. Additionally, we noticed that with higher sampling sizes, the proportion of samples undergoing temperature increases is higher compared to low sampling sizes, which aligns with our analysis presented in Section \ref{sec:diversity}.

\paragraph{Proportion of Optimal Temperature Range}
We analyze the proportion of FSD instances that ultimately reach the dead zone. We consider reaching the dead zone as an indication that the sample operates within an optimal temperature range. As shown in Figure~\ref{fig:analysis_fsd_ratio}, dynamic temperature sampling results in a higher proportion of FSD instances entering the dead zone compared to fixed-temperature sampling, suggesting that our method enables better alignment for a larger number of samples.
\section{Related Work}

\paragraph{Self-Consistency}
Self-consistency \citep{SC}, also known as majority voting, is a significant method for effectively enhancing the reasoning performance of large language models (LLMs) within the context of chain-of-thought \citep{COT} settings. 
Research on this method primarily focuses on two aspects: First, the effectiveness of self-consistency is further improved through weighted majority voting \citep{li-etal-2023-making,TDG} or input diversity \citep{PTSC}. Additionally, some have extended self-consistency to open-domain generation \citep{wang-etal-2024-integrate,SCOG}, allowing its application beyond reasoning tasks. Second, some studies aim to reduce the cost of self-consistency without compromising performance, according to early stopping criteria about answer distributions \citep{ESC,aggarwal-etal-2023-lets}, difficulty \citep{DSC}, quality \citep{RASC} or consistency of reasoning paths \citep{PC}. \citet{chen-etal-2024-self-para} have employed a hybrid strategy combining sampling and greedy algorithms to reduce computational costs. Recently, theoretical analyses of voting strategies \citep{ISL,ESC} were provided, offering a theoretical foundation for the study of self-consistency. Our method offers a deeper viewpoint, revisiting self-consistency from the perspective of distributional dynamic alignment.

\paragraph{Diversity Control for Language Models}
Decoding strategy is a critical factor in controlling the diversity of language models. From the perspective of the probability distribution of generated tokens, temperature sampling \citep{temperaturesample} controls the sharpness of the distribution by adjusting the temperature. Existing research primarily focuses on diversity control within a single sampling process \citep{EDT,HoC,ADLPO,li-etal-2024-dynamic}. At the task level, \citet{renze-2024-effect} have examined the impact of temperature on the model's problem-solving capabilities. However, the influence of diversity control on self-consistency and the underlying mechanisms remain unexplored.

\section{Conclusion}
This work revisits self-consistency through the lens of dynamic distributional alignment, challenging the conventional view of passive convergence to a fixed answer distribution. We demonstrate that decoding temperature critically shapes both sampling behavior and the latent answer distribution itself, revealing a trade-off between diversity-driven exploration and finite-sample convergence. By introducing a confidence-aware mechanism that dynamically adjusts temperature based on real-time alignment with the distribution, we bridge this gap, enabling efficient synchronization between sampling dynamics and evolving answer distributions. Empirical results validate that this approach outperforms static strategies, achieving robust performance improvements without external resources. Our findings position self-consistency as an active alignment challenge, opening avenues for adaptive aggregation frameworks in reasoning tasks.
\section*{Limitations}
While our approach advances the understanding and application of self-consistency, several limitations remain:  
\begin{itemize}
    \item Task Scope: Experiments focus on mathematical reasoning tasks, thus generalization to broader domains (e.g., open-ended generation or multi-step decision-making) requires further validation.
    \item Optimal Temperature: The specific value of the optimal temperature when the sample size approaches infinity, and how it varies with factors such as the model and dataset, remains unexplored.
    \item  Decoding Strategy Interactions: The interplay between temperature modulation and other decoding techniques (e.g., top-k or top-p sampling) remains unexplored, potentially affecting broader applicability.
\end{itemize}

\section*{Ethics Statement}
All of the datasets used in this study were publicly available, and no annotators were employed for our data collection. We confirm that the datasets we used did not contain any harmful content and was consistent with their intended use (research). We have cited the datasets and relevant works used in this study.

\section*{Acknowledgments}
This work is supported by Beijing Natural Science Foundation (No.4222037, L181010).


\bibliography{anthology,custom}

\newpage
\appendix
\section{Proof of Theorem~\ref{the:sc}}
\label{app:proof}
\begin{prof}
Firstly, we need to introduce true labels into Definition \ref{def:sc}. As we are not concerned with the specific numerical values of the predicted and true answers, we map the set of predicted answers onto a sequence of natural numbers (in simple terms, we only need to know which of the i-th answers is the correct one). Consequently, we can establish the following partial order relation:
\begin{align}
&\left|\; \underset{i}{argmax}\; f^M(i) - \underset{i}{argmax}\; f^N(i) \;\right| \notag\\
&=\left|\; [\underset{i}{argmax}\; f^M(i)-gt_j] \right.\notag\\
& \left. \qquad \qquad - [\underset{i}{argmax}\; f^N(i)-gt_j] \;\right| \notag\\
& \geq \left|\; \mathbb{I}[\underset{i}{argmax}\; f^M(i)=gt_j] \right. \notag \\
& \left. \qquad \qquad- \mathbb{I}[\underset{i}{argmax}\; f^N(i)=gt_j] \;\right| 
\end{align}
Based on Definition \ref{def:sc}, we have:
\begin{align}
    &\left|\; \mathbb{I}[\underset{i}{argmax}\; f^M(i)=gt_j] \right. \notag \\
    & \left. \qquad \qquad - \mathbb{I}[\underset{i}{argmax}\; f^N(i)=gt_j] \;\right| < \epsilon
\label{eq:single_gt_sc}
\end{align}
Next, we introduce the dataset $D$ into Equation \ref{eq:single_gt_sc}:
\begin{align}
    &\frac{1}{|D|}\sum_{j\in D}\left|\; \mathbb{I}[\underset{i}{argmax}\; f^M(i)=gt_j] \right. \notag \\
    &\left. \qquad \qquad - \mathbb{I}[\underset{i}{argmax}\; f^N(i)=gt_j] \;\right| < \epsilon 
\end{align}
According to $|a+b| \leq |a|+|b|$, we have:
\begin{align}
    &\left|\; \frac{1}{|D|}\sum_{j\in D}\mathbb{I}[\underset{i}{argmax}\; f^M(i)=gt_j] \right. \notag \\
    &\left. \; - \frac{1}{|D|}\sum_{j\in D}\mathbb{I}[\underset{i}{argmax}\; f^N(i)=gt_j] \;\right| < \epsilon 
\end{align}
Finally, we can derive Theorem \ref{the:sc}:
\begin{equation}
    \left| \; Acc_{D}^M - Acc_{D}^N \;\right| < \epsilon
\end{equation}
\end{prof}

\clearpage

\section{Additional Results of Section~\ref{sec:diversity}}
\label{app:diversity}

\begin{figure}[ht]
\centering
\includegraphics[width=2\linewidth]{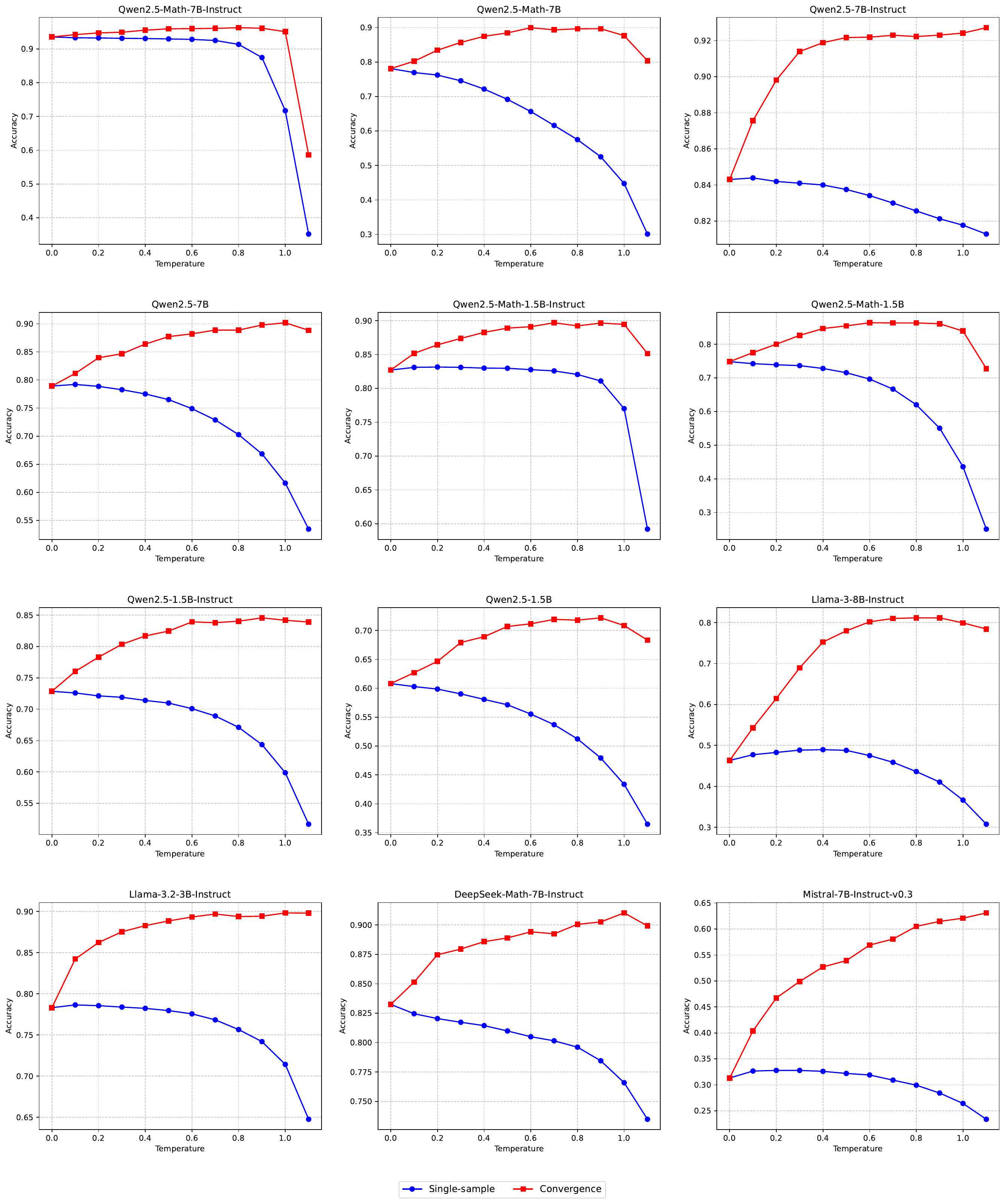}
\label{fig:inf_tem_all}
\end{figure}

\clearpage

\begin{figure}[ht]
\centering
\includegraphics[width=2\linewidth]{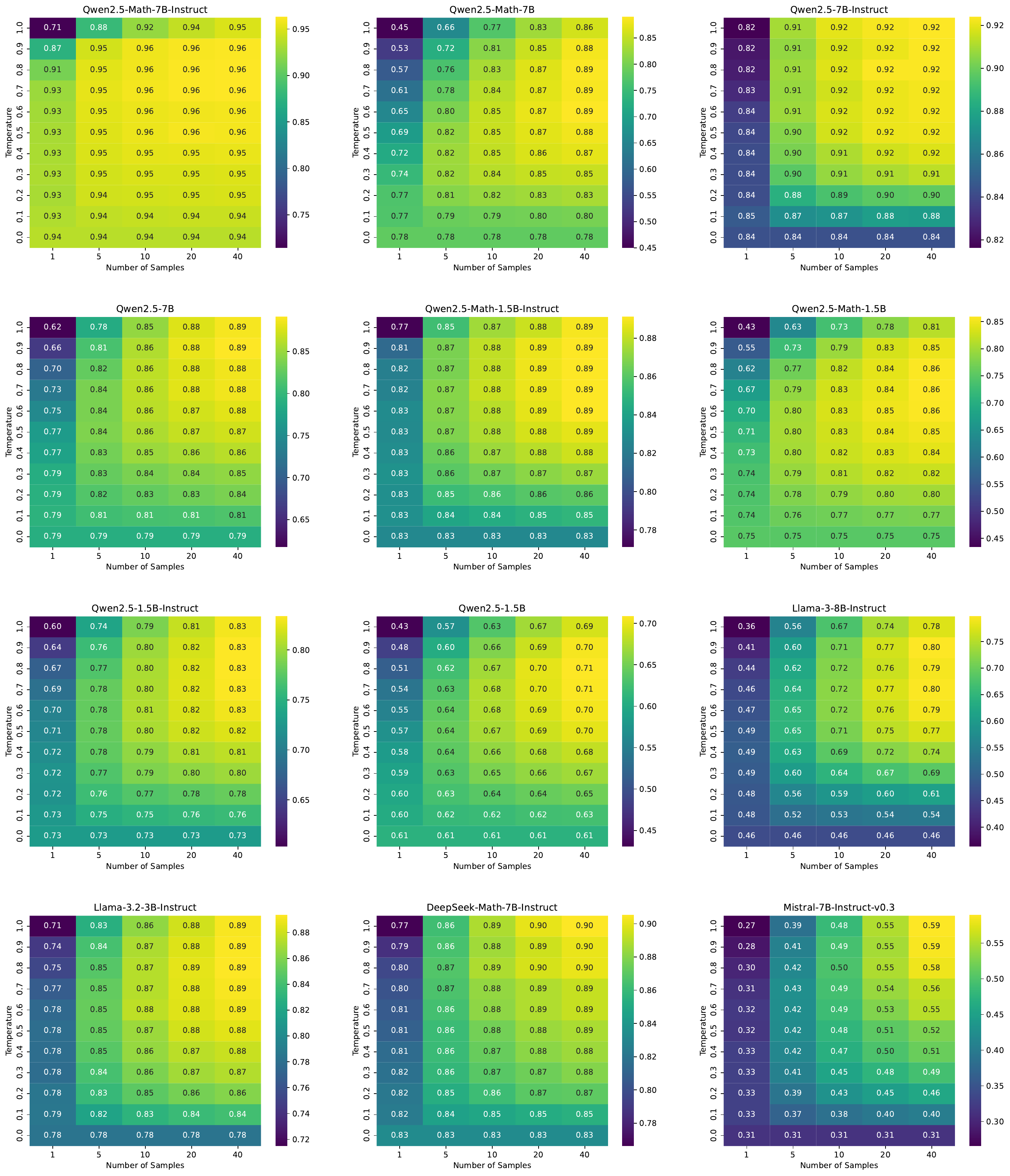}

\label{fig:lim_tem_all}
\end{figure} 
\clearpage

\section{Justification of the Top-2 Assumption}
To validate the correctness of the assumption in Section~\ref{sec:threshold} that the final answer can only appear among the top-2 answers,
we conducted an analysis to measure how often the final self-consistency answer overlaps with the top-2 answers observed in early samples.
\begin{table}[h]
  \centering
  \small
  \renewcommand\tabcolsep{3.5pt}
  \begin{tabular}{lccc}
    \toprule
    Model                           & $N = 10$ & $N = 20$ \\
    \midrule
    Qwen2.5-Math-7B-Instruct        & 99.44\%  & 99.73\%  \\
    Qwen2.5-Math-7B                 & 96.79\%  & 98.39\%  \\
    Qwen2.5-7B                      & 97.74\%  & 98.82\%  \\
    Qwen2.5-1.5B-Instruct           & 95.38\%  & 97.53\%  \\
    Qwen2.5-7B-Instruct             & 99.21\%  & 99.58\%  \\
    LLaMA-3-8B-Instruct             & 87.85\%  & 93.65\%  \\
    LLaMA-3.2-3B-Instruct           & 96.86\%  & 98.31\%  \\
    \bottomrule
  \end{tabular}
  \caption{Overlap Rate Between Top-2 and Final Answer.}
  \label{tb:top-2_justification}
\end{table}

The results in Table~\ref{tb:top-2_justification} suggest that the final majority-vote answer is overwhelmingly likely to appear among the top-2 candidates, even with a small number of samples. This supports the practical validity of the assumption.

\section{Additional Results of Section~\ref{sec:results}}
\begin{table*}[h]
    \centering
    \small
    \renewcommand\tabcolsep{3.5pt}
    \begin{tabular}{l l 
                    cc cc cc  
                    cc cc cc} 
    \toprule
    \multirow{3}{*}{Model} & \multirow{3}{*}{Strategy}
      & \multicolumn{6}{c}{Last Letter Concatenation}
      & \multicolumn{6}{c}{StrategyQA} \\
    & & \multicolumn{2}{c}{N=10} & \multicolumn{2}{c}{N=20} & \multicolumn{2}{c}{N=40}
      & \multicolumn{2}{c}{N=10} & \multicolumn{2}{c}{N=20} & \multicolumn{2}{c}{N=40} \\
    \cmidrule(lr){3-4}\cmidrule(lr){5-6}\cmidrule(lr){7-8}
    \cmidrule(lr){9-10}\cmidrule(lr){11-12}\cmidrule(lr){13-14}
    & & Mean & Max & Mean & Max & Mean & Max
      & Mean & Max & Mean & Max & Mean & Max \\
    \midrule
    \multirow{2}{*}{Qwen2.5-1.5B-Instruct}
      & Fix     & 12.7 & 16.6 & 15.6 & 19.5 & 18.4 & 22.2
                & 55.1 & 58.9 & 57.4 & 60.5 & 58.7 & 61.2 \\
      & Dynamic & \textbf{13.7} & \textbf{16.8} & \textbf{18.5} & \textbf{20.9} & \textbf{21.5} & \textbf{23.3}
                & \textbf{55.8} & \textbf{59.1} & \textbf{58.2} & \textbf{60.7} & \textbf{59.5} & \textbf{61.4} \\
    \midrule
    \multirow{2}{*}{LLaMA-3.2-3B-Instruct}
      & Fix     & 72.8 & 76.1 & 75.7 & 78.9 & 76.7 & 80.2
                & 67.1 & \textbf{70.5} & 67.9 & \textbf{71.3} & 68.3 & \textbf{71.5} \\
      & Dynamic & \textbf{73.4} & \textbf{76.5} & \textbf{76.6} & \textbf{79.4} & \textbf{78.2} & \textbf{81.0}
                & \textbf{68.2} & \textbf{70.5} & \textbf{68.8} & \textbf{71.3} & \textbf{68.9} & \textbf{71.5} \\
    \midrule
    \multirow{2}{*}{Mistral-7B-Instruct-v0.3}
      & Fix     & 4.8  & 5.7  & 6.0  & 7.9  & \textbf{6.8}  & 9.5
                & 52.3 & 55.0 & 55.2 & 59.9 & 56.6 & 62.3 \\
      & Dynamic & \textbf{5.1}  & \textbf{6.1}  & \textbf{6.2}  & \textbf{8.9}  & \textbf{6.8}  & \textbf{10.3}
                & \textbf{52.6} & \textbf{55.9} & \textbf{55.8} & \textbf{60.7} & \textbf{57.5} & \textbf{62.9} \\
    \bottomrule
    \end{tabular}
    \caption{Evaluation results on Last Letter Concatenation and StrategyQA tasks.}
    \label{tb:addition-results}
\end{table*}

The results in Table~\ref{tb:addition-results} demonstrate that our method performs well across reasoning tasks in different domains.

\section{Additional Analysis of Section~\ref{sec:analysis}}

\subsection{Influence of Model Architecture and Calibration Properties}
The underlying model architecture can influence the effectiveness of self-consistency, particularly due to differences in reasoning ability, calibration behavior, and sensitivity to temperature. To explore this, we conducted a comparative analysis across different backbone models, using several indicators:
(1) Confidence is measured via answer entropy and FSD.
(2) Stability is measured via the variance (Var) of accuracy under different fixed temperatures.
(3) Effectiveness is reflected by both the absolute accuracy under fixed-temperature self-consistency (Acc fix@N) and the performance gains brought by our adaptive method (Gain@N).

\begin{table*}[h]
  \centering
  \small
  \renewcommand\tabcolsep{3.5pt}
  \begin{tabular}{lccccccccc}
    \toprule
    Model                         & Entropy & FSD  & Var  & Acc@10 & Gain@10 & Acc@20 & Gain@20 & Acc@40 & Gain@40 \\
    \midrule
    Qwen2.5-1.5B-Instruct         & 1.97                 & 0.483          & 4.089            & 79.0   & +0.2    & 80.3   & +0.5    & 81.1   & +0.5    \\
    Qwen2.5-7B-Instruct           & 1.17                 & 0.656          & 2.120            & 90.8   & +0.0    & 91.2   & +0.2    & 91.4   & +0.3    \\
    LLaMA-3.2-3B-Instruct         & 1.96                 & 0.526          & 2.146            & 86.2   & +0.0    & 87.2   & +0.3    & 87.7   & +0.4    \\
    LLaMA-3-8B-Instruct           & 2.61                 & 0.349          & 52.387           & 66.6   & +0.5    & 70.2   & +1.4    & 76.1   & +1.9    \\
    \bottomrule
  \end{tabular}
  \caption{Comparison of model uncertainty, stability, and effectiveness under fixed-temperature and adaptive method.}
  \label{tb:uncertainty-accuracy-gains}
\end{table*}

According to Table~\ref{tb:uncertainty-accuracy-gains}, our key observation is that higher confidence models (lower entropy, higher FSD) tend to:
(1) achieve higher base accuracy under fixed temperature, and
(2) exhibit lower sensitivity to temperature (i.e. lower variance), resulting in smaller performance gains from adaptive strategies. These trends align well with our understanding of model behavior: stronger models tend to produce more confident predictions, making them inherently less reliant on temperature-based sampling adjustments. Conversely, less confident models benefit more from dynamic temperature calibration, as their sampling distributions are more sensitive to the choice of temperature.

\subsection{Relationship Between Sample Difficulty and Temperature Variation}
To better understand the reasons behind the differing behavior of the observed temperature variation between samples, we hypothesize that sample difficulty is a key prior factor. Intuitively, harder questions tend to result in lower model confidence and may require lower initial temperatures to guide convergence, whereas easier questions are more stable and better explored with higher temperatures.

To examine this hypothesis, we conducted further analysis using samples with known or estimated difficulty levels. For the MATH dataset, we use its ground-truth difficulty labels (1–5). For GSM8K, we used an LLM-based difficulty estimation strategy, where we applied repeated batch-wise comparisons to assign continuous difficulty scores (1–8).

For each initial temperature $T_0$, we group the samples based on whether the final temperature increases, decreases, or remains the same, and then calculate the average difficulty score within each group.

\begin{table*}[h]
  \centering
  \small
  \renewcommand\tabcolsep{3.5pt}
  \begin{tabular}{c  ccc  ccc}
    \toprule
    \multirow{2}{*}{$T_0$}
      & \multicolumn{3}{c}{GSM8K} 
      & \multicolumn{3}{c}{MATH} \\
    \cmidrule(lr){2-4} \cmidrule(lr){5-7}
      & $\uparrow$ $T_3$ Avg Level 
      & $\rightarrow$ $T_3$ Avg Level 
      & $\downarrow$ $T_3$ Avg Level 
      & $\uparrow$ $T_3$ Avg Level 
      & $\rightarrow$ $T_3$ Avg Level 
      & $\downarrow$ $T_3$ Avg Level \\
    \midrule
    0.2 & 4.41 & 5.27 & 6.06  & 3.29 & 4.08 & 4.21 \\
    0.3 & 4.43 & 5.34 & 5.90  & 3.19 & 4.05 & 4.16 \\
    0.4 & 4.24 & 5.44 & 5.63  & 3.03 & 3.95 & 4.10 \\
    0.5 & 4.22 & 5.27 & 5.67  & 3.00 & 3.86 & 4.10 \\
    0.6 & 4.20 & 5.25 & 5.70  & 2.99 & 3.81 & 4.10 \\
    0.7 & 4.20 & 5.24 & 5.65  & 2.98 & 3.78 & 4.09 \\
    0.8 & 4.16 & 5.20 & 5.61  & 2.94 & 3.71 & 4.10 \\
    0.9 & 4.12 & 5.09 & 5.66  & 2.91 & 3.73 & 4.04 \\
    \bottomrule
  \end{tabular}
  \caption{Average sample difficulty levels by temperature adaptation results across different initial temperatures for GSM8K and MATH.}
  \label{tb:sample_difficulty}
\end{table*}

The results in Table~\ref{tb:sample_difficulty} show a clear and consistent pattern:
(1) More difficult questions tend to lead to temperature decreases, while easier questions often allow for temperature increases.
(2) Higher initial temperatures generally result in more samples decreasing their temperature during the adaptive process.
This aligns well with our intuition and further supports the idea that the final temperature $T_3$ is implicitly influenced by sample difficulty.

\end{document}